\documentclass[9pt]{extarticle}
\usepackage{spconf,amsmath,graphicx, multirow, booktabs, comment, tabularx, xcolor, url, subcaption, makecell}

\title{Towards End-to-End Spoken Grammatical Error Correction}
\name{Stefano Bannò, Rao Ma, Mengjie Qian, Kate M. Knill, Mark J.F. Gales\thanks{This paper reports on research supported by Cambridge University Press \& Assessment (CUP\&A). Mengjie Qian is supported by EPSRC Project EP/V006223/1 (Multimodal Video Search by Examples).}}

\address{ALTA Institute, Department of Engineering, University of Cambridge, UK}
\begin{document}
%
\maketitle
\begin{abstract}
Grammatical feedback is crucial for L2 learners, teachers, and testers. Spoken grammatical error correction (GEC) aims to supply feedback to L2 learners on their use of grammar when speaking. This process usually relies on a cascaded pipeline comprising an ASR system, disfluency removal, and GEC, with the associated concern of propagating errors between these individual modules. In this paper, we introduce an alternative ``end-to-end'' approach to spoken GEC, exploiting a speech recognition foundation model, Whisper. This foundation model can be used to replace the whole framework or part of it, e.g., ASR and disfluency removal. These end-to-end approaches are compared to more standard cascaded approaches on the data obtained from a free-speaking spoken language assessment test, Linguaskill. Results demonstrate that end-to-end spoken GEC is possible within this architecture, but the lack of available data limits current performance compared to a system using large quantities of text-based GEC data. Conversely, end-to-end disfluency detection and removal, which is easier for the attention-based  Whisper to learn, does outperform cascaded approaches. Additionally, the paper discusses the challenges of providing feedback to candidates when using end-to-end systems for spoken GEC.
\end{abstract}
\section{Introduction}
\label{sec:intro}
In natural language processing and, specifically, computer-assisted language learning, the task of grammatical error correction (GEC) has been a topic of significant interest and research. Traditionally, this task primarily focused on the correction of written text, such as essays, documents, or emails, to enhance their grammatical accuracy and fluency. This is a well established area of study~\cite{bryant2022grammatical}, with four shared tasks organised in the last 15 years.
As interest in automating all skills in language learning, including speech, has increased, there is a need to apply GEC to a broader range of data. Spoken GEC tackles the complex challenge of correcting grammatical errors within spoken language. Unlike written text, spoken language presents unique characteristics, including disfluencies, hesitations, truncated words and sentences, and a lack of punctuation and capitalisation. These factors make the task of correcting spoken grammatical errors considerably more complicated and fascinating. So far, there has been a limited number of studies that have investigated spoken grammar using automated methods. This exploration began with the pioneering work by \cite{izumi2003automatic}, which involved manual transcriptions of Japanese learners of English. In recent years, there has been an emergence of fully automated approaches in this domain~\cite{Knill2019, caines2020, lu2020spoken, lu2022onassessing}. A major constraint on progress in this research area has been the limited availability of specifically designed and annotated data. Motivated by this lack of data, spoken GEC is typically structured as a sequential process comprising three distinct modules. First, an automatic speech recognition (ASR) module is employed to transcribe the spoken content. Subsequently, a module for disfluency detection (DD) and removal comes into play, tasked with removing disfluencies such as interruptions, repetitions, and hesitations from the spoken discourse. Finally, a GEC system tuned to handle speech transcriptions is used to correct grammatical errors.

In this paper, we propose the use of a speech recognition foundation model - Whisper - to perform end-to-end spoken GEC and DD. Exploiting Whisper for end-to-end spoken language understanding has been recently proposed~\cite{wang2023whislu}, but, to the best of our knowledge, it has never been investigated for DD or GEC. In Section \ref{sec:proposed_method}, we present the end-to-end and cascaded systems. Metrics and evaluation methods are discussed in Section \ref{sec:evaluation_methods}. Section \ref{sec:experimental_results} is the core part of this paper: after outlining data and model setups, we describe our experiments. The first block illustrates DD experiments conducted on a publicly available data set, Switchboard~\cite{holliman1992switchboard, calhoun2010nxt, zayats2019disfluencies}, while the second block is devoted to experiments on both DD and GEC on an L2 learner data set obtained from the Speaking module of Linguaskill~\cite{ludlow2020official}, a medium-stake language test. Subsequently, we discuss our findings in relation with their implications on learner feedback, focusing on the challenges posed by end-to-end systems. Finally, in Section \ref{sec:conclusions}, we outline the conclusions and highlight the potential avenues for further research.

\section{Proposed Method}
\label{sec:proposed_method}

\subsection{End-to-end System}
\label{sec:e2e}

  
  

Recently, foundation ASR models built from large-scale datasets have been released. One popular model - Whisper \cite{radford2022robust} - is trained on more than 680 thousand hours of labelled data covering 97 languages and shows good performance on a wide range of standard data sets without fine-tuning. In training, several tasks are jointly learned by the model including ASR, speech translation, voice activity detection, and language identification. Since the model is trained on large quantities of data in a multi-task fashion, we assume it has the ability to perform tasks that require general language understanding other than plain ASR. In this work, we propose to adapt Whisper to generate outputs in the desired format for different target tasks. The model is tuned separately on three types of manual references to achieve effective adaptation: \textbf{1)} original ASR transcriptions with disfluencies labelled (\textbf{dsf}, Whisper$_\text{dsf}$), \textbf{ 2)} fluent transcriptions with hesitations, false starts, etc. removed (\textbf{flt}, Whisper$_\text{flt}$), and \textbf{3)} grammatically corrected transcriptions (\textbf{gec}, Whisper$_\text{gec}$).

\begin{figure}[!htbp]
    \centering
    \includegraphics[width=1\linewidth]{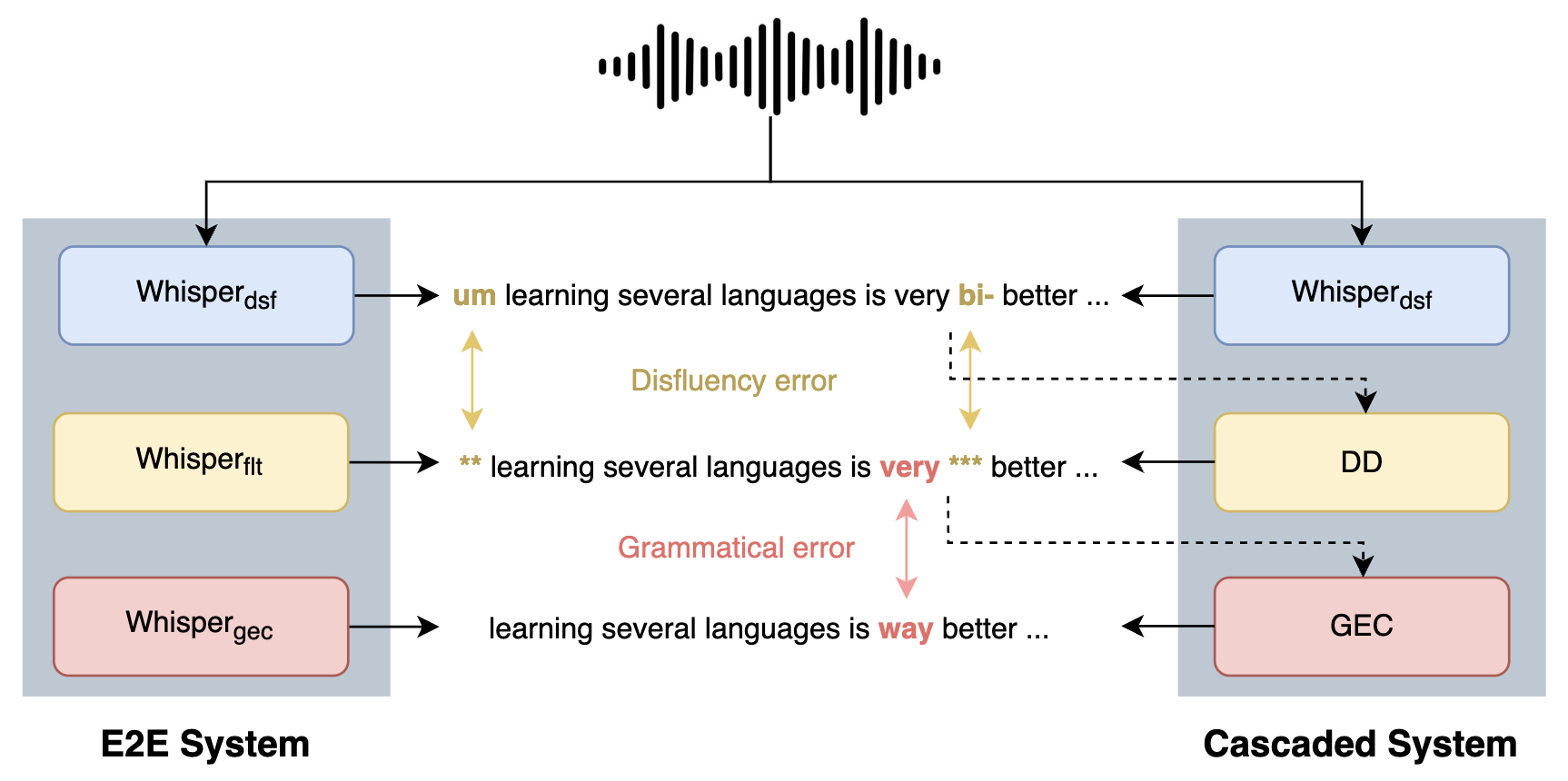}
    \caption{Illustration of an E2E SGEC system and a cascaded system.}
    \label{fig:sgec}
\end{figure}

Two types of task adaptation methods are explored: fine-tuning (FT) and soft prompt tuning (SPT) \cite{lester2021power,ma23_slate}. In FT, all the model parameters are updated. Instead, for SPT, a small amount of continuous vectors are inserted into the decoder embedding space and tuned on the training set while keeping the original model parameters fixed. Here, 20 soft prompt vectors with a dimension size of 768 are learned for each task, accounting for only 0.006\% of parameters compared to FT. Thus, SPT is far more parameter-efficient allowing the same core foundation model to be used for multiple tasks. For a given utterance, the adapted foundation ASR models can be used to generate outputs for three individual tasks, as illustrated on the left-hand side of Figure \ref{fig:sgec}.

\subsection{Cascaded System}

In contrast to end-to-end approaches, a traditional cascaded spoken GEC system comprises separate ASR, DD, and GEC modules, as shown on the right-hand side of Figure \ref{fig:sgec}.

\noindent \textbf{ASR:} the ASR system used is the Whisper$_\text{dsf}$ model described in Section \ref{sec:e2e}, which is trained on the dsf manual references.

\noindent \textbf{DD:} we approached the task of DD by employing a BERT-based token classifier for sequence tagging:
\[\textrm{\textbf{d}}_{1:M} = \textrm{BERT}(w_{1:M}) 
\quad 
p(r_{m}|w_{1:M}) = f_{d}(\textrm{\textbf{d}}_{m})
\]
where 
$\textbf{d}_{m} $ is the word embedding associated to word \emph{w\textsubscript{m}},
\emph{r\textsubscript{m}} is a binary tag which indicates whether word \emph{w\textsubscript{m}} is fluent or disfluent; $f_{d}$ is trained to estimate the probability that a given word \emph{w\textsubscript{m}} is fluent.
Subsequently, we eliminated all words identified as disfluencies from the transcriptions. In particular, the BERT-based model comprises the following components: a pre-trained BERT model sourced from the HuggingFace Transformer Library~\cite{huggingface} (\emph{bert-base-uncased}), a dropout layer, a dense layer with 768 nodes, an additional dropout layer, a second dense layer with 128 nodes, and, finally, the output layer.

\noindent \textbf{GEC:} we perform GEC as a sequence-to-sequence task. For the GEC model, we used a BART model~\cite{bart2020} initialised from the version provided by the HuggingFace Transformer Library \cite{huggingface} (\emph{facebook/bart-base}).\footnote{Preliminary experiments included GECToR~\cite{gector2020}, but we achieved better performances using BART.}

\section{Evaluation Methods}
\label{sec:evaluation_methods}

While it is straightforward to use word error rate (WER) to evaluate ASR performance, assessing DD and spoken GEC is more challenging when an ASR system is used to generate the transcriptions~\cite{lu2022onassessing}.

\noindent \textbf{DD:} is assessed using Precision, Recall, and F$_\text{1}$ scores, but this requires annotation of individual words. This is problematic when using multiple ASR systems and, hence, different decoded transcriptions. Therefore, we use WER as our primary metric to evaluate DD~\cite{lu2022onassessing} and only consider Precision, Recall, and F$_\text{1}$ scores as additional metrics for feedback analysis (see Sections \ref{sec:swbd} and \ref{sec:feedback}).

\noindent \textbf{GEC:} numerous metrics have been devised for assessing written GEC. The General Language Evaluation Understanding (GLEU)~\cite{napoles2015ground} score, inspired by BLEU~\cite{papineni2002bleu}, employs n-gram precision relative to a reference. It rewards both word-level corrections and faithfully preserved source text. On the other hand, MaxMatch $M^2$~\cite{dahlmeier2012better} captures phrase-level edits and computes F$_\text{0.5}$ scores accordingly. This metric is particularly suitable for feedback-oriented applications that focus on edits. However, as we have already observed in our previous work~\cite{lu2022onassessing}, when it comes to evaluating spoken GEC, applying these standard metrics is not straightforward. A common challenge in cascaded-style spoken language applications is the difficulty of comparison across systems when upstream modules are different. For example, the input text to the GEC module varies when the upstream ASR and DD models change. If we were to apply GLEU and $M^2$ scores in such cases, these metrics would provide different results every time the ASR transcriptions change. Consequently, the results are not comparable across systems. Moreover, it is crucial to remember that even for end-to-end trained spoken systems, evaluation metrics are not clearly defined. Transferring written-based metrics to spoken tasks is challenging since end-to-end systems do not yield any intermediate variables for assessment. For this reason, we adopt both WER and translation edit rate (TER)~\cite{snover2006study} as the primary evaluation metrics for GEC, while we employ Precision, Recall, and F$_\text{0.5}$ scores as additional metrics in the feedback analysis section (see Section \ref{sec:feedback}).

\section{Experimental Results}
\label{sec:experimental_results}

\subsection{Dataset}

\noindent \textbf{Switchboard:} the Switchboard corpus consists of manual transcriptions of 260 hours of telephone conversations by first language (L1) American English speakers~\cite{holliman1992switchboard}. 
For our experiments in Section \ref{sec:swbd}, we used the NXT version~\cite{calhoun2010nxt}, which contains disfluency annotations and time-alignment information. This allowed us to work using the audio recordings and the respective transcriptions consistently. We only removed the sentences containing the $\tt{MUMBLEX}$ token, used by human annotators in case of unintelligible words, and those only consisting of filler words (``uh-hum'', etc.). A version of the corpus has been recently expanded and reannotated with better-quality disfluency annotations~\cite{zayats2019disfluencies}. As it does not contain time-alignment information, we could only use this version in training the cascaded text-based DD model in Section \ref{sec:linguaskill}.
In our experiments, we treat annotated disfluencies and hesitations/filler words (such as “uh”, “uhm”, etc.) as disfluencies. For training/dev/test partition, we refer to the guidelines indicated in \cite{charniak2001edit}. Further information about the data can be found in Table \ref{T:data_stats}.

\noindent \textbf{Linguaskill:} the data used in our study are obtained from candidate responses to the Speaking module of the Linguaskill tests for L2 learners of English, provided by Cambridge University Press \& Assessment~\cite{ludlow2020official}. The data set is balanced for gender and features around 30 L1s and proficiency levels ranging from A2 to C of the Common European Framework of Reference (CEFR) \cite{cefr2001}. It has been manually tagged with annotations about disfluencies and grammatical error corrections~\cite{knill23_slate}. Responses can be up to 60 seconds in length, so here they were split into `sentences' using automatic time alignment of manually marked boundaries between speech phrases. The reader can refer to Table \ref{T:data_stats} for further information about the data.

\begin{table}[!htbp]
\footnotesize
    \centering
    \begin{tabular}{l|l|c|c|c|c|c}
    \toprule
    & Corpus &  Split & Hours & Speakers & Utts/Sents & Words \\
    \midrule
    \multirow{6}*{\rotatebox{90}{Spoken}} & \multirow{3}*{Switchboard} & train & 50.8 & 980 & 81,812 & 626K \\
    & & dev & 3.8 & 102 & 5,093 & 46K \\
    & & test & 3.7 & 100 & 5,067 & 45K \\
    \cmidrule{2-7}
    & \multirow{3}*{Linguaskill} & train & 77.6 & 1,908 & 34,790 & 502K \\
    & & dev & 7.8 & 176 & 3,347 & 49K \\
    & & test & 11.0 & 271 & 4,565 & 69K \\
    \midrule
    \multirow{3}*{\rotatebox{90}{Written}} & \multirowcell{3}{EFCAMDAT\\+BEA-2019} & \multirowcell{3}{train\\dev} & \multirowcell{3}{-\\-} & \multirowcell{3}{-\\-} & \multirowcell{3}{2.5M\\25,529} & \multirowcell{3}{28.9M\\293K} \\
    &&&&&&\\
    &&&&&&\\
    \bottomrule
    \end{tabular}
    \caption{Statistics of datasets.}
    \label{T:data_stats}
\end{table}

\noindent \textbf{EFCAMDAT+BEA-2019:} arguably, the largest publicly available L2 learner corpus, the second release of the EF-Cambridge Open Language Database (EFCAMDAT) \cite{geertzen2013automatic} comprises 1,180,310 assignments written by 174,743 L2 learners. The L1s of the learners are not available, but can be inferred from their nationalities (about 200). Furthermore, the learner scripts are annotated with proficiency scores, part-of-speech tags, and information on grammatical dependencies, and are partially corrected by human experts. 
As the data set contains noisy responses and incorrect annotations, we used only part of the responses. The data cleaning process is described in \cite{banno23b_slate}.

A shared task on GEC was organised within the Workshop on Innovative Use of NLP for Building Educational Applications~\cite{bryant2019bea}. The organisers released a collection of written corpora tagged with GEC annotations\footnote{More information can be found here: \url{cl.cam.ac.uk/research/nl/bea2019st/#data}} that we used in our experiments. Punctuation and capitalisation have been removed from both the EFCAMDAT and BEA-2019 data to make them more similar to speech transcriptions. Further information can be found in Table \ref{T:data_stats}.

\subsection{Model Setup}

\noindent \textbf{Whisper:} there are multiple sizes of Whisper models released and the small.en model is used as the foundation model in this paper. The pre-trained model is tuned on the Linguaskill training set with different manual references for three tasks as per Section \ref{sec:e2e}. With both FT and SPT adaptation methods, the model is trained for 30,000 steps on the training set. A batch size of 5 is used in the training. The learning rates of FT and SPT are initialised to 1e-5 and 0.1 separately, and linear decay is applied in the training. In the decoding of Whisper, beam search with a width of 5 is adopted.

\noindent \textbf{DD:} the model employed in the first part of our experiments (Section \ref{sec:swbd}) is trained on the Switchboard NXT data~\cite{calhoun2010nxt} for 4 epochs with maximum sequence length 128, batch size 64, dropout rate 0.2, and learning rate 5e-6.
For the other experiments, we trained the model on the reannotated version of Switchboard~\cite{zayats2019disfluencies} for 5 epochs with the same parameters as before. It was further fine-tuned on the Linguaskill data for 8 epochs with learning rate 2e-6.

\noindent \textbf{GEC:} the BART model was trained on the EFCAMDAT and BEA-2019 data for 19 epochs with maximum sequence length 256, batch size 16, gradient accumulation step 4, and learning rate 2e-6. It was further fine-tuned on the Linguaskill data for 5 epochs with the encoder frozen, and the learning rate is reduced to 1e-5.

\subsection{Results on Switchboard}
\label{sec:swbd}
Initially, performance is evaluated on the publicly available data, Switchboard. Table \ref{T:whisper_vs_man_swbd} shows the system performance in terms of WER against both the disfluent (original) and fluent manual references. Whisper$_\text{dsf}$ and Whisper$_\text{flt}$ indicate Whisper trained on disfluent and fluent transcriptions, respectively. Although our main aim is not speech recognition, it is still interesting to evaluate performance of Whisper$_\text{dsf}$ on the standard, disfluent, references as this allows comparison with the published results~\cite{radford2022robust}. However, of more interest within the context of this work is the performance of Whisper$_\text{flt}$ for end-to-end ASR and DD. This outperforms the cascaded DD approach, as shown in Table \ref{T:whisper_vs_man_swbd}.

\begin{table}[!htbp]
    
    \centering
    \begin{tabular}{l|c|c}
    \toprule
        \multicolumn{1}{c|}{Model} & dsf & flt \\
        
        \midrule
        Whisper$_\text{dsf}$ & 10.62 & 14.83 \\
        
        Whisper$_\text{dsf}$+DD & - & 10.86 \\
        
        Whisper$_\text{flt}$ & 13.83 & \textbf{10.32} \\
    \bottomrule
    \end{tabular}
    \caption{Evaluation of Whisper FT performance against disfuent (dsf) and fluent (flt) manual references on Switchboard in terms of \%WER.}
    \label{T:whisper_vs_man_swbd}
\end{table}

DD is typically evaluated using Precision, Recall, and F$_\text{1}$ scores. It is interesting to examine how this metric could be examined for end-to-end systems. Here, we calculate scores based on the deletions between the fluent ASR transcriptions from Whisper$_\text{flt}$ and the disfluent transcriptions of Whisper$_\text{dsf}$. This can be compared with the cascaded approach and the automatic disfluent transcriptions, as shown in Table \ref{T:p_r_f1_swbd}. Note these performance figures require that a word is correctly identified and tagged for deletion as specified in the manual transcriptions and disfluency labels.

\begin{table}[!htbp]
    \centering
    \begin{tabular}{c|c|c|c}
    \toprule
        DD Model &  P & R & F\textsubscript{1}\\ \hline
        Whisper$_\text{flt}$ $\xrightarrow{{\tt del}}$ Whisper$_\text{dsf}$ & 49.82   & 55.00    & 50.52   \\
        Whisper$_\text{dsf}$+DD $\xrightarrow{{\tt del}}$ Whisper$_\text{dsf}$ & 69.18    & 68.21    & 67.21   \\

    \bottomrule
    \end{tabular}
    \caption{Evaluation of DD in terms of Precision, Recall, and F$_\text{1}$ scores on the Switchboard test set based on deletions.}
    \label{T:p_r_f1_swbd}
\end{table}

The reasons for the discrepancy between these results and the WER performance shown in Table \ref{T:whisper_vs_man_swbd} can be attributed to the fact that the cascaded system compares deletions based on a single transcription, whereas two different transcriptions produced with two different decoding processes are contrasted for the end-to-end system. Hence, disfluencies in the end-to-end system can result from transcription differences as well as deletions in the fluent system from removing disfluencies. More in-depth analysis will be presented in Section \ref{sec:feedback}.

\subsection{Results on Linguaskill}
\label{sec:linguaskill}

The main focus of this work is related to spoken GEC. Here, we use transcribed Linguaskill data~\cite{ludlow2020official} as described in Table \ref{T:data_stats} to evaluate end-to-end DD and GEC. The results in Table \ref{T:whisper_vs_man_linguaskill} show that Whisper can be tuned to yield both end-to-end DD and GEC, with the best WER performance in each case coming from the matched system. This is true for fine-tuning (FT) and soft prompt tuning (SPT). The former consistently outperforms the latter, so for further experiments, we focus on the end-to-end systems built using FT. The SPT model performance, however, is reasonable, with drop of only 0.3-0.8\% WER. Given the reduction in parameters to train and store, SPT has practical advantages over FT.

\begin{table}[!htbp]
    \centering
    \small
    \begin{tabular}{l|cc|cc|cc}
    \toprule
        \multicolumn{1}{c|}{\multirow{2}*{Model}} & \multicolumn{2}{c|}{dsf} & \multicolumn{2}{c|}{flt} & \multicolumn{2}{c}{gec}\\
        & FT & SPT & FT & SPT& FT & SPT\\
        \midrule
        Whisper$_\text{dsf}$ & \textbf{5.92} & 6.36 & 9.97 & 10.58 & 19.17 & 19.58 \\
        Whisper$_\text{dsf}$+DD  & ---  & --- & 6.31 & 6.82 & --- & --- \\
        Whisper$_\text{flt}$ & 9.22 & 9.55 & \textbf{5.77} & 6.34 & 14.89 & 15.29 \\
        Whisper$_\text{gec}$ & 13.73 & 12.51 & 10.37 & 9.26 & \textbf{13.49} & 14.22 \\
    \bottomrule
    \end{tabular}
    \caption{Evaluation of Whisper FT and SPT performance against different manual references on Linguaskill in terms of \%WER.}
    \label{T:whisper_vs_man_linguaskill}
\end{table}

As seen on Switchboard, the end-to-end DD system, Whisper$_\text{flt}$, achieves a lower WER than applying a DD specific model to the disfluent Whisper output (5.77\% vs 6.31\%).

For spoken GEC, the end-to-end approach, Whisper$_\text{gec}$, is compared to a standard cascaded system, Whisper$_\text{dsf}$+DD+GEC, and a cascaded system combining Whisper$_\text{flt}$ and GEC. Though the end-to-end performance in Table \ref{gec_performance} is comparable to the standard cascaded system (TER of 13.08\% vs 12.96\%), the best-performing model is the cascaded system exploiting the Whisper$_\text{flt}$ model. This result must be taken in the context that the GEC module used in the cascaded approaches is trained on a large amount of text-based GEC training data and has been additionally fine-tuned on Linguaskill transcriptions. On the other hand, Whisper$_\text{gec}$ was only trained on the Linguaskill data. Even leveraging far less GEC training data, the end-to-end Whisper$_\text{gec}$ model achieves comparable performance to a traditional cascaded system.

\begin{table}[htbp!]
    
    \centering
    \begin{tabular}{l|l|c|c }
        \toprule
         \multirow{2}*{System} &  \multicolumn{1}{c|}{\multirow{2}*{ASR Model}} &  \multicolumn{2}{c}{gec} \\
         & & WER & TER \\
        
        \midrule
        \multirow{2}*{Baseline} 
         & Whisper$_{\text{dsf}}$ & 19.17  & 18.74 \\
         
         & Whisper$_{\text{flt}}$ & 14.89  & 14.49  \\

        \midrule
        \multirow{2}*{Cascaded} 
         & Whisper$_{\text{dsf}}$+DD+GEC  & 13.34   & 12.96    \\
         
         & Whisper$_{\text{flt}}$+GEC & \textbf{12.96}   & \textbf{12.54}    \\

        \midrule
        E2E & Whisper$_{\text{gec}}$ & 13.49  & 13.08 \\
        
        \bottomrule
    \end{tabular}
    \caption{GEC results with FT baseline, cascaded, and E2E systems.}
    \label{gec_performance}
\end{table}

A closer look to the WER breakdown reported in Table \ref{wer_breakdown} shows that Whisper$_{\text{flt}}$ is mainly performing deletions when compared to the manual disfluent transcriptions, in line with what we would expect from a model for disfluency removal. Conversely, when compared to the manual fluent transcriptions, Whisper$_{\text{gec}}$ mostly performs substitutions and, to a lesser degree, insertions and deletions.

\begin{table}[!htbp]
    
    \centering
    \begin{tabular}{c|c|c|c}
    \toprule
        \multirow{2}*{Model} & \multicolumn{3}{c}{WER (Sub/Del/Ins)} \\
        & dsf & flt & gec \\
        \midrule
        Whisper$_\text{dsf}$ & 3.3/1.3/1.4 & 3.2/0.8/5.9 & 8.2/3.2/7.7 \\
        
        \midrule
        Whisper$_\text{flt}$ & 2.9/5.4/0.9 & 3.0/1.3/1.5 & 7.8/3.8/3.4 \\

        \midrule
        Whisper$_\text{gec}$ & 5.3/6.2/2.3 & 5.4/2.1/2.9 & 6.8/3.2/3.5 \\

    \bottomrule
    \end{tabular}
    \caption{WER breakdown (Sub/Del/Ins) of Whisper transcription against different manual references.}
    \label{wer_breakdown}
\end{table}

Table \ref{T:example_1} illustrates an example drawn from the data. As can be observed, a notorious issue with the evaluation of GEC performance is that there are often multiple potentially correct solutions (e.g., in this case \emph{way} and \emph{much}), but only one of them is generally contemplated for scoring purposes. Here, the correction of \emph{way} in Whisper$_\text{gec}$ output is considered as an error compared to the manual reference while also leading to a grammatically correct result.

\begin{table}[!htbp]
    \centering
    \setlength{\tabcolsep}{4pt} 
    \renewcommand{\arraystretch}{1.2} 
    \begin{tabularx}{\columnwidth}{l|X}
    \toprule
        Type & Sentence \\
        
        \midrule
        Ref$_\text{flt}$ & actually learning several languages is very better than just learn one language because it's more easy to talk with people from all around the world \\
        \midrule
        Whisper$_\text{flt}$ & actually learning several languages is very better than just learn one language because it's more easy to talk with people from all around the world \\
        \midrule
        Ref$_\text{gec}$ & actually learning several languages is \textcolor{orange}{much} better than just \textcolor{orange}{learning} one language because it's \textcolor{orange}{easier} to talk with people from all around the world \\
        \midrule
        Whisper$_\text{gec}$ & actually learning several languages is \textcolor{orange}{way} better than just \textcolor{orange}{learning} one language because it's \textcolor{orange}{easier} to talk with people from all around the world \\
    \bottomrule
    \end{tabularx}
    \caption{Example of a Whisper$_\text{gec}$ transcription. Corrections in orange.}
    \label{tab:my_label}
    \label{T:example_1}
\end{table}

\subsection{Feedback analysis}
\label{sec:feedback}

An important aspect of helping learners improve their spoken language and inform teachers is fine-grained feedback of where and how a learner has been disfluent or made grammatical errors. To this end, it is not sufficient to just produce the fluent or grammatically corrected transcriptions.

As in Section \ref{sec:swbd}, we can evaluate DD on the Linguaskill data in terms of Precision, Recall, and F$_\text{1}$ scores based on deletions between Whisper$_\text{flt}$ against the Whisper$_\text{dsf}$ transcripts.
\begin{table}[!htbp]
\small
    \centering
    \begin{tabular}{c|c|c|c|c}
    \toprule
        DD Model & Strategy &  P & R & F\textsubscript{1}\\
        \midrule
        \multirow{2}*{Whisper$_\text{flt}$ $\xrightarrow{{\tt del}}$ Whisper$_\text{dsf}$} & SPT  & 66.63  & 70.97   & 66.84   \\
        & FT  & 61.02  & 68.11   & 62.30  \\
        \midrule
        \multirow{2}*{Whisper$_\text{dsf}$+DD $\xrightarrow{{\tt del}}$ Whisper$_\text{dsf}$} & SPT  & 74.95   & 75.00   & 73.28  \\
       & FT  & 74.94   & 75.05   & 73.35  \\

    \bottomrule
    \end{tabular}
    \caption{Evaluation of DD in terms of Precision, Recall, and F$_\text{1}$ scores on the Linguaskill test set based on deletions.}
    \label{T:p_r_f1_linguaskill}
\end{table}
Similarly to what we observed in our previous experiments, we find a discrepancy between the performance in terms of Precision, Recall, and F$_\text{1}$ scores reported in Table \ref{T:p_r_f1_linguaskill} and the WER performance shown in Table \ref{T:whisper_vs_man_linguaskill}. As stated above, this can be explained by considering that the cascaded system simply uses a tagger that labels one set of transcriptions, whereas the end-to-end system employs two different sets of transcriptions which are obtained through two different decoding processes. The better Precision, Recall, and F$_\text{1}$ scores achieved using SPT clearly support this explanation.

\begin{table}[!htbp]
    \small
    \centering
    \begin{tabular}{c|c|c|c}
    \toprule
        \multirow{2}*{Model} & \multirow{2}*{Strategy}  & \multicolumn{2}{c}{Whisper$_\text{dsf}$} \\
        && WER & Sub/Del/Ins \\
        \midrule
        \multirow{2}*{Whisper$_\text{flt}$} & SPT & 12.70  & 1.2/11.3/0.2  \\
        & FT & 13.89  & 2.1/11.5/0.3  \\
        \midrule
        \multirow{2}*{Whisper$_\text{dsf}$+DD} & SPT & 11.34  & 0.0/11.3/0.0  \\
        & FT & 11.34  & 0.0/11.3/0.0  \\
    \bottomrule
    \end{tabular}
    \caption{Overall WER and breakdown (Sub/Del/Ins) of Whisper$_\text{flt}$ and Whisper$_\text{dsf}$+DD against Whisper$_\text{dsf}$ hypotheses.}
    \label{T:wer_breakdown_flt_vs_dsf}
\end{table}

Furthermore, Table \ref{T:wer_breakdown_flt_vs_dsf} offers more insights in favour of our hypothesis. As can be observed, while in the cascaded systems, WER only consists of deletions, the comparison between Whisper$_\text{flt}$ and Whisper$_\text{dsf}$ also involves a modest number of substitutions and a marginal number of insertions. Also, once again, we see that the SPT results seem to corroborate our theory in that substitutions are halved and insertions are also reduced.

Proceeding in a similar fashion for GEC, we evaluate performance in terms of Precision, Recall, and F$_\text{0.5}$ scores by considering GEC edit labels between the Whisper$_\text{flt}$ transcripts and the Whisper$_\text{gec}$ ones. These are extracted using the ERRor ANnotation Toolkit (ERRANT)~\cite{bryant2017automatic}, which is a standard approach in GEC analysis and evaluation. Examples of ERRANT edit labels are \texttt{R:VERB:FORM}, which indicates an incorrect verb form, and \texttt{M:DET}, which indicates a missing determiner. ERRANT also labels error types as \texttt{OTHER} when edits do not fall under any other category. A large part of errors labelled as \texttt{OTHER} are paraphrases. The prefix \texttt{R:} stands for \emph{replace}, \texttt{M:} for \emph{missing}, and \texttt{U:} for \emph{unnecessary}. Since the off-the-shelf version of ERRANT operates at a span-based level, we modified it in a way that our hypothesis and reference edits are aligned to account for ASR insertions/deletions. We did this by shifting the span of hypothesis edits to the left in the case of
a deletion and to the right in the case of an insertion.

\begin{table}[!htbp]
\small
    \centering
    \begin{tabular}{c|c|c|c|c}
    \toprule
        GEC Model & Strategy &  P & R & F\textsubscript{0.5}\\
        \midrule
        \multirow{2}*{Whisper$_\text{gec}$ $\xrightarrow{{\tt gec}}$ Whisper$_\text{flt}$} & SPT  & 25.41   & 14.60     & 22.13    \\
        & FT  & 27.77   & 22.31   & 26.40   \\
        \midrule
        \multirow{2}*{Whisper$_\text{flt}$+GEC $\xrightarrow{{\tt gec}}$ Whisper$_\text{flt}$}  & SPT  & 43.53    & 25.91    & 38.31   \\
       & FT  & 44.70    & 27.53    & 39.74  \\ \hline \hline
       Manual$_\text{flt}$+GEC $\xrightarrow{{\tt gec}}$ Manual$_\text{flt}$ & -  & 58.64    & 35.84   & 52.02 \\

    \bottomrule
    \end{tabular}
    \caption{Evaluation of GEC in terms of Precision, Recall, and F$_\text{0.5}$ scores on the Linguaskill test set based on GEC edits.}
    \label{T:gec_analysis_linguaskill}
\end{table}

Unlike what we observed while analysing the DD performance, the results for the GEC models are in line with their performance in terms of WER and TER reported in Table \ref{gec_performance}. The cascaded system including Whisper$_\text{flt}$ and the text-based GEC model also shows better results in terms of Precision, Recall, and F\textsubscript{0.5} scores.

\begin{figure}[!htbp]
    \centering
    \includegraphics[width=1\linewidth]{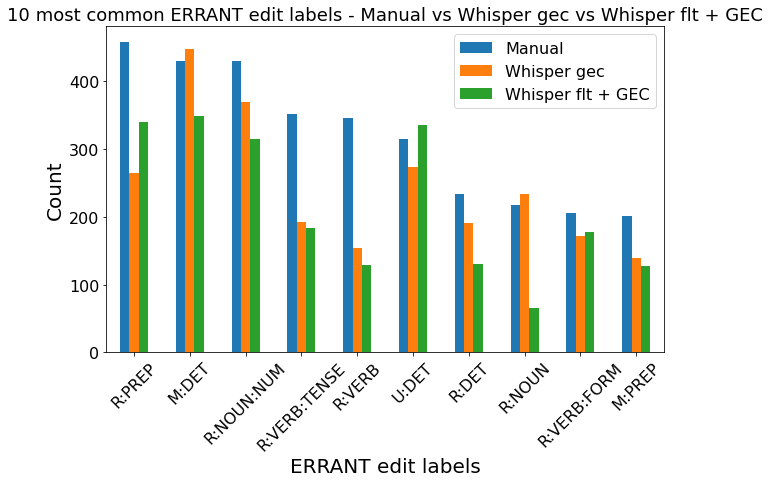}
    \caption{10 most common ERRANT edit labels.}
    \label{ERRANT_edits_man_whisper}
\end{figure}



Another interesting piece of analysis stems from the analysis of the ERRANT edit labels in the manual references, the end-to-end system, and the partially cascaded system (see Figure \ref{ERRANT_edits_man_whisper}). Out of the 10 most common ERRANT edit labels (excluding the \texttt{OTHER} labels), 9 are the same when comparing the manual references to the end-to-end system (as \texttt{M:PREP} is the 11th most common in the end-to-end system). Therefore, end-to-end spoken GEC is able to detect most of the common forms of grammatical error.

Table \ref{gec_performance_individual} reports the GEC evaluation results in terms of Precision, Recall, and F$_\text{0.5}$ scores considering all the 9 most common ERRANT edit labels individually for the end-to-end and cascaded GEC systems. The need for more training data for the end-to-end system is particularly evident when we consider \texttt{R:VERB} and \texttt{R:NOUN}. These labels indicate errors related to word usage and require the word spoken to be replaced with a different word. An example drawn from the data is the following:

\begin{quote}
    \item \emph{in my opinion if we have a problems with quality of this product we can attend more staff to solve this problems but not for constant time}
\end{quote}

In this case, apart from the article errors and the inappropriate use of the adjective \emph{constant}, the verb \emph{attend} should be replaced with \emph{hire}. On the other hand, the results on other edit labels are more in line with the cascaded system, such as \texttt{R:VERB:FORM}, for which we report this example drawn from the data:

\begin{quote}
    \item \emph{about my hometown i like to going out with my dogs spending time in the countryside}
\end{quote}

in which \emph{going} should be corrected into \emph{go}.

This suggests that this type of switch may be observed in the Whisper training data.

\begin{table}[htbp!]
    
    \centering
    \small
    \begin{tabular}{l|c c c|c c c}
        \toprule
         \multirow{2}*{Error} &  \multicolumn{3}{c}{W$_\text{gec}$ $\xrightarrow{{\tt gec}}$ W$_\text{flt}$} &  \multicolumn{3}{c}{W$_\text{flt}$+GEC $\xrightarrow{{\tt gec}}$ W$_\text{flt}$} \\
         & P & R & F$_\text{0.5}$ & P & R & F$_\text{0.5}$ \\
        
        \midrule
        \texttt{R:PREP} & 54.45  & 31.81  & 47.66  & 54.77  & 40.54  & 51.18  \\
        \texttt{M:DET} & 39.69  & 40.96  & 39.94  & 54.70  & 43.93  & 52.14  \\
        \texttt{R:NOUN:NUM} & 46.96  & 40.18  & 45.43  & 61.20  & 43.79  & 56.69  \\
        \texttt{R:VERB:TENSE} & 35.39  & 19.87  & 30.61  & 45.96  & 23.34  & 38.50  \\
        \texttt{R:VERB} & 10.74  & 4.70  & 8.55  & 42.06  & 15.59  & 31.40  \\
        \texttt{U:DET} & 41.33  & 34.78  & 39.83  & 45.21  & 46.89 & 45.54  \\
        \texttt{R:DET} & 22.58  & 18.26  & 21.56  & 41.54  & 23.48  & 36.00  \\
        \texttt{R:NOUN} & 5.53  & 5.69 & 5.56  & 36.21  & 9.95  & 23.70  \\
        \texttt{R:VERB:FORM} & 44.25  & 40.10  & 43.35  & 50.29  & 44.79  & 49.06  \\

        \bottomrule
    \end{tabular}
    \caption{Evaluation of GEC in terms of Precision, Recall, and F$_\text{0.5}$ scores on the Linguaskill test set focusing on individual GEC edits.}
    \label{gec_performance_individual}
\end{table}

\section{Conclusions and future work}
\label{sec:conclusions}
In this work, we have investigated an end-to-end approach to spoken GEC. While our best-performing model is the cascaded system that applies GEC to Whisper fine-tuned on fluent transcriptions (i.e., Whisper$_\text{flt}$), it is noteworthy that the end-to-end system (i.e., Whisper$_\text{gec}$) achieves comparable performance to a conventional cascaded system. This is particularly interesting given that the foundation ASR model was fine-tuned on only a limited quantity of data. However, one of the challenges for end-to-end systems is feedback for learners. End-to-end spoken GEC systems by definition yield only the grammatically correct speech,  whereas cascaded systems are able to yield grammar edits. Though edits can be obtained by comparing two end-to-end systems, fluent speech versus grammatically correct speech, this limits the quality of the feedback. As part of our future work, we intend to explore alternative foundational models and conduct a more extensive analysis of feedback options.

\section{Acknowledgements}
Thanks to CUP\&A for access to the Linguaskill Speaking data and the ELiT Annotation Team led by Diane Nicholls for the annotations of that data. Special thanks to Thomas Hardman and Vyas Raina for providing the modified version of ERRANT.

\bibliographystyle{IEEEbib}
\bibliography{strings,refs}

\end{document}